\pdfoutput=1

\documentclass[11pt]{article}

\usepackage{acl}

\usepackage{times}
\usepackage{latexsym}

\usepackage[T1]{fontenc}

\usepackage[utf8]{inputenc}

\usepackage{microtype}

\usepackage{inconsolata}

\usepackage{hyperref}
\usepackage{url}
\usepackage{booktabs, multirow, graphicx}
\usepackage{subcaption}

 \usepackage{amsmath} 
\usepackage{algorithm}
\usepackage{algpseudocode}
\usepackage{wrapfig}
\usepackage{adjustbox}
\usepackage{xspace}
\usepackage{xcolor}

\newcommand{\flan}{Flan-T5\xspace}
\newcommand{\llama}{LLaMA\xspace}
\newcommand{\ours}{CoBa\xspace}
\newcommand{\ourslong}{Correction with Backtracking\xspace}
\newcommand{\newsroom}{Newsroom\xspace}
\newcommand{\cnn}{CNN/Dailymail\xspace}
\newcommand{\xsum}{XSUM\xspace}
\title{Correction with Backtracking  Reduces Hallucination
in Summarization}

\author{Zhenzhen Liu\thanks{$^\ast$Part of the work is conducted during an internship at Google.} \quad Chao Wan \quad Varsha Kishore \quad Jin Peng Zhou\\
  Cornell University\\
  \texttt{\{zl535, cw862, vk352, jz563\}@cornell.edu} \\
  \AND
  Minmin Chen\\
  Google DeepMind\\
  \texttt{minminc@google.com} \\\And
  Kilian Q. Weinberger\\
  Cornell University\\
  \texttt{kqw4@cornell.edu}}

\begin{document}
\maketitle
\begin{abstract}
Abstractive summarization aims at generating natural language summaries of a source document that are succinct while preserving the important elements. Despite recent advances, neural text summarization models are known to be susceptible to hallucinating (or more correctly confabulating), that is to produce summaries with details that are not grounded in the source document. In this paper, we introduce a simple yet efficient technique, \ours{}, to reduce hallucination in abstractive summarization. The approach is based on two steps: hallucination detection and mitigation. We show that the former can be achieved through measuring simple statistics about conditional word probabilities and distance to context words. Further, we demonstrate that straight-forward backtracking is surprisingly effective at mitigation. We thoroughly evaluate the proposed method with prior art on three benchmark datasets for text summarization. The results show that \ours{} is effective and efficient in reducing hallucination, and offers great adaptability and flexibility. Code can be found at \url{https://github.com/zhenzhel/CoBa}.
\end{abstract}

\section{Introduction}

Recent summarization methods, based on neural sequence-to-sequence and language models (LM), are able to produce high-quality summaries~\citep{pegasus, flant5, llama}.   
However, despite their impressive capabilities 
these summarization models are prone to hallucinations, a phenomenon where models make statements that seem plausible but are not grounded in the source document~\citep{pagnoni2021understanding, maynez2020faithfulness,zhao2020reducing}. Hallucinations compromise the accuracy and trustworthiness of the generated summaries. 

\begin{figure}[ht!]
    \centering
    \includegraphics[width=\linewidth]{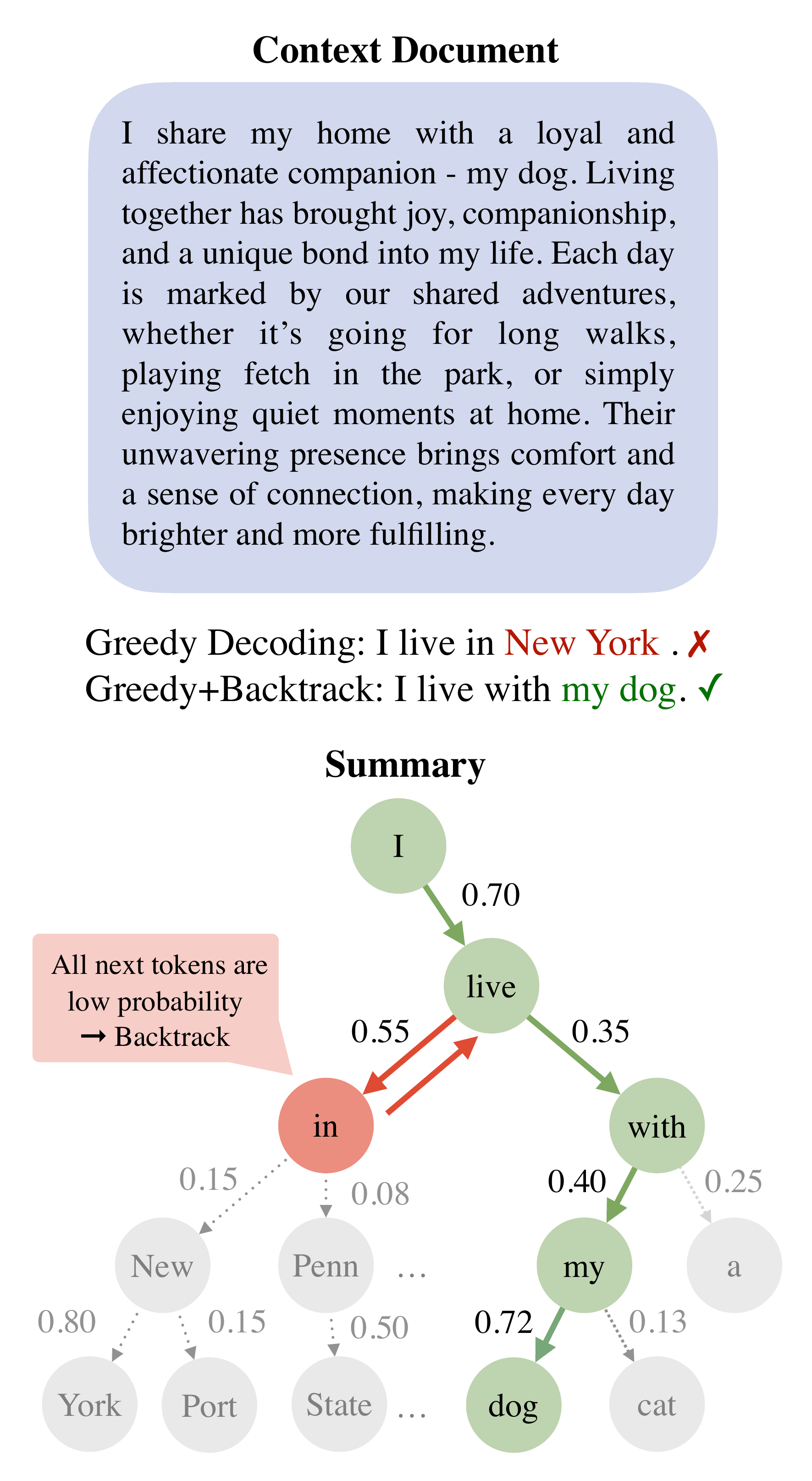}
    \caption{Schematic illustration of \ours{} (using only token probability as the detection metric with threshold $0.2$). After the partial summary \textit{``I live''}, the token \textit{``in''} has a higher probability than ``with". However, \textit{``I live in''} will pressure the model into hallucinating a place. We detect this because all the next tokens have a probability lower than our threshold $0.2$. Backtracking enables the model to find an alternative continuation that avoids hallucination down the line.}
    \label{fig:main}
\end{figure}

We hypothesize that one reason for hallucination is that sometimes after a LM generates partial text, there is no completion that is grounded in the source text. An illustration of this situation is shown in \autoref{fig:main}. Although the partial sentence \emph{I live in} is highly plausible, it forces the LM to specify where the person lives, even though this is not specified in the source document.  
Such situations can often be \emph{detected} by intrinsic properties of hallucinated text: (1) the first word of a  hallucinated sequence tends to have low conditional probability, (2) hallucinations are not supported by words in the context, and therefore have a large distance to context words. 
Returning to our previous example, if the language model continues the sentence \textit{I live in} without any support from the context, \textit{Munich} might be just as plausible as \textit{New York}, or \textit{Penn State}. None of the locations would have particularly high probability, therefore triggering condition (1). Further, if none of the cities are mentioned in the context, all would have a large word distances to the context words, triggering condition (2). 
Once the beginning of a hallucination is detected, we 
\textit{backtrack} and re-generate the \emph{preceding}   words that ``cornered'' the LM into a position without a faithful continuation. In our example, we replace the token \textit{in} by the token \textit{with}; consequently, based on the context, the generated sentence can be completed with \textit{my dog}.  

Our method \emph{\ourslong (\ours)}, is a simple inference-time method that requires no additional model training and is compatible with most of the decoding methods. We evaluate \ours{} on three established document summarization datasets and measure the faithfulness of generated summaries. We show that it is highly effective and efficient for detecting and mitigating hallucinations. \ours is also orthogonal to many existing hallucination reduction techniques and can be used in conjunction to those.

\section{Background and Related Work}\label{sec:related}

We adopt the definition of hallucination for abstractive summarization from \citet{maynez2020faithfulness}: 
\textit{The summary $\mathcal{S}$ for a context document $\mathcal{C}$ contains hallucinations if there exists a span in $\mathcal{S}$ which is not supported by $\mathcal{C}$.}

Hallucinations  exhibit task-specific characteristics in various Natural Language Generation (NLG) tasks. For instance, in Machine Translation, hallucination is often observed in the output when the input source undergoes specific perturbation~\citep{lee2018hallucinations}. In Question Answering (QA), one common manifestation is semantic drift, where the generated answers deviate from the topic of the question~\citep{li2021addressing}. Additionally, in retrieval-based QA, the retrieval model may introduce additional sources of hallucination~\citep{ji2023survey}.  

Various existing works seek to understand how hallucination happens, and have identified several factors. In various datasets, human generated ground truth summaries can contain additional information not present in the corresponding input texts~\citep{dhingra2019handling,wang2020asking}. Training on such data may increase a model's tendency to hallucinate. 
During generation, hallucination may occur when the model attends to irrelevant parts of the input context~\citep{tian2019sticking}, or utilizes knowledge acquired during training that is not grounded in the context~\citep{longpre2021entity}. Additionally, the decoding method also impacts the faithfulness of generation. Past work has observed that sampling-based decoding can lead to increased hallucination~\citep{dziri2021neural, lee2022factuality, lookahead}.

\subsection{Methods for Reducing Hallucination}
Depending on the task and problem setup, various methods have been developed to detect and mitigate hallucinations. 
Existing approaches can be broadly categorized into training time mitigation and generation time mitigation.  

\paragraph{Training Time Mitigation.}Noise in the pre-training corpus is shown to be a significant source of hallucination for language models~\citep{zhou2023lima}. Some past work has focused on applying simple mechanisms to filter training data, many of which are already used in training large language models~\citep{touvron2023llama, penedo2023refinedweb, li2023textbooks}. Data curation is not only done in the pre-training stage but also can happen during supervised finetuning (SFT). 
Researches in this area focus on using high-quality, human curated, or domain-specific data~\citep{elaraby2023halo} for SFT and have shown that this can lead to improved faithfulness~\citep{zhou2023lima, chen2023alpagasus, lee2023platypus, cao2023instruction}.

\paragraph{Generation Time Mitigation.} Recent publications have also explored how to enhance the faithfulness of generation during inference time~\citep{zhang2023siren}. One line of work performs post-editing by training specialized models~\citep{cao2020factual, chen2021improving, dong2020multi} or by directly prompting the models~\citep{varshney2023stitch, mundler2023self}. Others modify the decoding algorithm.
\citet{lee2022factuality} proposes to gradually decrease the value of $p$ in top-$p$ sampling (i.e. nucleus sampling), to reduce hallucinations introduced by randomness. 
\citet{li2023inference} modifies attention to encourage more factual generations. \citet{shi2023trusting} proposes \emph{Context-Aware Decoding (CAD)} to suppress hallucinations arising from the model's prior knowledge; they adjust the context-conditional token logits with the unconditional logits. \citet{lookahead} proposes \emph{Lookahead}: At each decoding step, it rolls out future summaries for the top $k$ tokens with the highest probabilities, adjusts their probabilities with BS-Fact, and picks the token with the highest adjusted probability. 
They also show that the performance can be further improved by ranking multiple candidates with a composite faithfulness score, or by distilling student models with the generated summaries. In contrast to these methods, \ours{} does not tamper with token probabilities. Instead, it detects hallucinated tokens and fixes them through backtracking and local edits (see \autoref{fig:main}). 

Most similar to our work is arguably \citet{king2022don}, a publication that we were not aware of until after the completion of this paper. While we do have distinct design choices and evaluations, we acknowledge that the two methods are rather similar and expect them to perform similarly under our setting.

\section{Problem Setup}

\begin{figure}[ht!]
    \captionsetup{font=small}
    \adjustbox{valign=t}{
    \begin{minipage}{\linewidth}    \includegraphics[width=\linewidth]{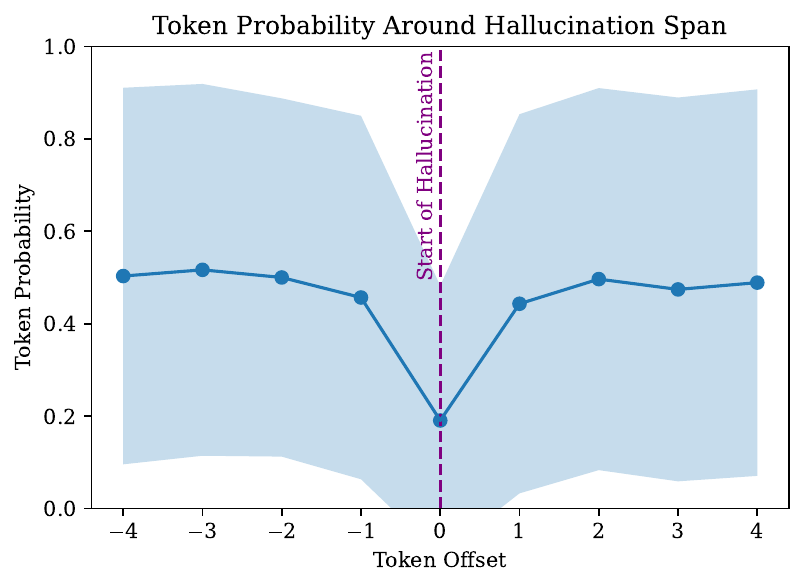}
    \end{minipage}}
    \adjustbox{valign=t}{
    \begin{minipage}{\linewidth}
        \centering
\includegraphics[width=\linewidth]{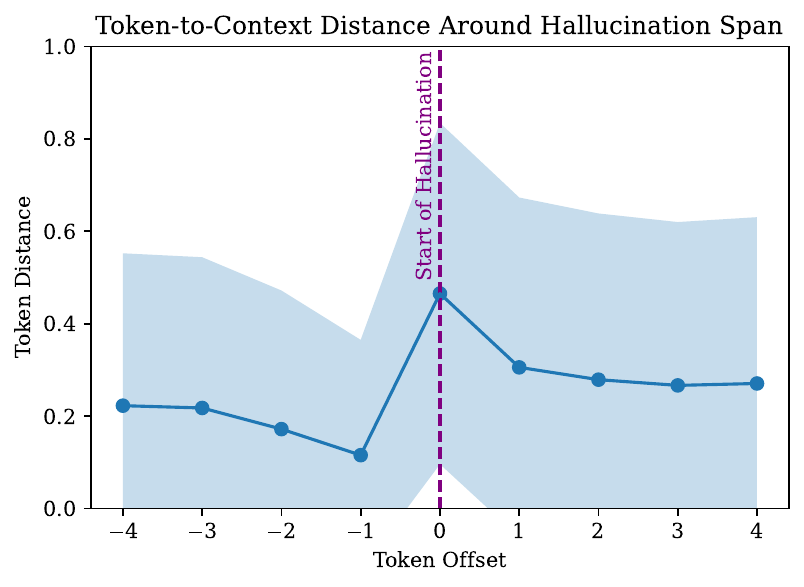}
\caption{\textbf{Average token probability (top) and token-to-context distance (bottom) around the hallucination span.} Token offset 0 stands for the token where hallucination starts, negative offsets stand for the tokens before hallucination and positive ones are for the hallucinated tokens. On average, the token which starts the hallucination has the lowest probability and is the furthest away from the context tokens compared to surrounding ones.}
\label{fig:dist}
\label{fig:confidence}
    \end{minipage}}
\end{figure}
Let $\mathcal{M}_{\theta}$ be an autoregressive summarization model with parameters $\theta$, and let $\Sigma$ be its vocabulary. Given a context document $\mathcal{C} = (c_1, \cdots c_m)$ as input, $M_\theta$ produces a summary $\mathcal{S} = (s_1,  \cdots, s_{n})$: 
\begin{equation}
\mathcal{M}_{\theta}(\mathcal{C}) = \mathcal{S}\nonumber
\end{equation}
where $c_1, \cdots c_m, s_1, \cdots, s_n \in \Sigma$; $m$ and $n$ are the lengths of the context and the summary respectively. In practice, $\mathcal{M}_{\theta}$ can either be a specialized summarization model like PEGASUS~\citep{pegasus}, or a general language model capable of zero-shot summarization like \flan~\citep{flant5}. %
If $\mathcal{M}_{\theta}$ requires prompting, we add a prompt like ``summarize: " to the context as input. 

Model $\mathcal{M}_{\theta}$ generates the summary autoregressively. At each step, given a partially generated summary $\mathcal{S}_{<t}$ up to token $s_{t-1}$, it outputs a distribution $
    p_{\theta} (s_t|\mathcal{C},\mathcal{S}_{<t})
$ for the next token $s_t$ over the vocabulary $\Sigma$. The probability of generating the summary $\mathcal{S}$ is thus 
\begin{equation*}
    p(\mathcal{S}) = \prod_{t=1}^{|\mathcal{S}|}p_{\theta} (s_t|\mathcal{C},\mathcal{S}_{<t})
\end{equation*}

\section{Reducing Hallucination at Inference}
We present a detection-correction approach for reducing hallucination at decoding time. The main idea is illustrated in \autoref{fig:main}: 
If a hallucination occurs, the problem typically originates already with its preceding tokens. The partially decoded summary can ``corner" the model such that there is no faithful next token. For example, in \autoref{fig:main}, the natural continuation for the partial summary ``I live in" is a name of a place. The source context however does not mention any places. We design strategies to detect such occurrences, and use backtracking~\citep{tarjan1972depth} to find alternative phrases that prevent hallucinations down the line.

\subsection{Hallucination Detection} \label{sec:detection}
We investigate different properties of hallucinated text and devise two strategies for detecting text that is not grounded in the context.

\subsubsection{Uncertainty-based Detection}

The intuition behind uncertainty-based detection is that hallucination is likely to occur if the model is unsure about what it should generate next conditioning on the input. 
The conditional probability of a token is one way of measuring uncertainty and prior work has shown that the token-wise probability of autoregressive language models is well-calibrated~\citep{kadavath2022language}. \citet{petryk2023caption} also use a similar technique for evaluating and ranking the correctness of image captions.

We validate that token probabilities are effective for identifying hallucinated tokens in summaries by computing probabilities on an  annotated hallucination dataset from \citet{googletaxonomy}. The dataset contains generated summaries from different summarization models, such as finetuned BERT~\citep{devlin2018bert}, Pointer-Generator Model~\citep{ptgen} and several more, with human annotations for hallucination spans. \autoref{fig:confidence} presents the conditional token probabilities of \flan XL around the hallucination span. Offset~0 represents where the hallucination starts, the negative offsets represent preceding tokens and the positive offsets represent successive tokens. In the figure, we observe a significant drop in token confidence at the start of hallucination. The average probability is only 0.2 in contrast with 0.5-0.6 for non-hallucinated tokens. The distribution of the probabilities is noisy shown as wide standard deviation in the figure, because of annotation noise and because some generated summaries can contain unnatural segments. 

Therefore, measuring conditional token probability is one way of detecting the beginning of hallucinations during the decoding process, when all possible next tokens have low probability, it suggests the absence of a suitable candidate, and potentially signals the onset of hallucination. Formally, at step $t$, we flag the token if the following condition holds:
\begin{equation*}\label{eq:prob}
    p_{\theta}(s_t | \mathcal{C}, \mathcal{S}_{<t}) < \delta
\end{equation*}
where %
$\mathcal{C}$ is the context document, $\mathcal{S}_{<t}$ is the partially generated summary, and $\delta$ is the token level conditional probability threshold for hallucination.

\subsubsection{Similarity-based Detection}
Another intuitive way of detecting hallucination is to find tokens in the generated summary that are not supported by the context, i.e., tokens that are not ``close" to any part of the context document. One method of measuring closeness is by computing cosine distance in the embedding space of a language model. More concretely, given a proposed token, we compute the distance between its embedding and the embeddings of all tokens in the context and flag the token as a potential hallucination if the minimum distance is above a certain threshold. The detection criterion in this case is:
\begin{equation}\label{eq:dist}
   d(v, \mathcal{C}) = \min_{c_i \in \mathcal{C}} \textit{cos\_dist}\big (\text{Emb}(v), \text{Emb}(c_i) \big) > \varphi \nonumber
\end{equation}
where $v$ is the proposed token, $\mathcal{C}$ is the context document and $\varphi$ is the distance threshold. \autoref{fig:dist} presents the minimum token-to-context distance computed over the annotated dataset from \citet{googletaxonomy}'s with embeddings from Flan-t5 XL (the results are averaged over 5000 samples). The average token distance at the first word in a hallucination span is significantly higher than words at other positions, as expected.

\subsection{Hallucination Mitigation}\label{sec:mitigation}
After detecting potential hallucination during decoding using the techniques described in \autoref{sec:detection}, we perform a local intervention to prevent the generation of hallucinated phrases. Specifically, we introduce a process similar to depth first search. We eliminate the last generated token $s_t$ and try to propose an alternative token $s_t'$ that does \textit{not} satisfy the hallucination criteria. We keep track of the eliminations given a partial sequence $\mathcal{S}_{<t}$ and context $\mathcal{C}$ to avoid repetitive proposals. If $s_t'$ can be found we add it to the generation and continue the forward decoding. We also continue if the partial sequence $\mathcal{S}_{<t}$ only contains the start-of-sequence token \texttt{[SOS]}. Otherwise, we backtrack again, i.e. eliminate the current last token $s_{t-1}$ and repeat the process (see \autoref{fig:main} for a pictorial description). 

Admittedly, sometimes the model is unable to find a good solution, and this is signaled by backtracking too many times. We therefore introduce an upper bound $L$, for the number of decoding steps (both forward and backtracking) that can be performed. We pick $L = 10 T$ where $T$ is the maximum generation length for our model $\mathcal{M}_\theta$. If an acceptable summary cannot be generated in $L$ steps, we turn off the backtracking mechanism and adopt greedy decoding to generate the summary. We empirically observe that with reasonable threshold choices, less than 3\% of the generations exceed the upper bound $L$ when using moderate threshold values in general.

\section{Experiments}
\begin{table*}[ht!]
\caption{\textbf{Faithfulness of the summaries generated with various decoding methods using \flan.} All the metrics are computed between the context document and the generated summary; higher is better.}
\label{tab:flan}
\begin{center}
\resizebox{0.7\linewidth}{!}{%
\begin{tabular}{c|lcccc}
\toprule
\textbf{} & \textbf{Method} &\textbf{AlignScore$\uparrow$} & \textbf{FactCC$\uparrow$}  & \textbf{BS-Fact$\uparrow$} & \textbf{Rouge-L$\uparrow$}   \\
\midrule
\multirow{16}{*}{\rotatebox{90}{Newsroom}} &Greedy   & 0.765 &	 0.604 & 0.919  & 0.131  \\
&+ Lookahead (every 8 tok.) & 0.768 & 0.607  & 0.920 & 0.133  \\
&+ Lookahead (every 4 tok.) & 0.774 & 0.607	  & 	0.922 & 0.136	 \\
&+ Lookahead (every 2 tok.) & 0.811 &  0.662  & 0.931 & 0.153  \\
&+ Lookahead (every tok.) & 0.816 & 0.662  & 0.933  & 0.159  \\
&+ CAD & 0.746 & 0.490  & 0.916  & 0.145 \\
&+ \ours & 0.821 & 0.674  & 0.923 & 0.138 \\
&+ \ours-d & 0.865 & 0.709  & 0.926 & 0.145\\
&+ \ours + CAD & 0.773 & 0.515  & 0.919 & 0.149 \\
&+ \ours-d + CAD & 0.820 & 0.560  & 0.922 & 0.161 \\
\cline{2-6}
&Nucleus & 0.636 &0.482  & 0.902 & 0.101 \\
&+ CAD & 0.694 & 0.430  & 0.907 & 0.117 \\
&+ \ours & 0.800 & 0.645  & 0.920 & 0.128 \\
&+ \ours-d & 0.857 &0.692  & 0.923 & 0.139\\
&+ \ours+ CAD & 0.767 & 0.505  &  0.917 & 0.139\\
&+ \ours-d + CAD &  0.817 & 0.552  & 0.921& 0.154 \\
\hline
\multirow{16}{*}{\rotatebox{90}{\xsum}} & Greedy & 0.723 & 0.485  & 0.919 & 0.096\\
&+ Lookahead (every 8 tok.) & 0.727 & 0.486   & 0.919 &  0.096 \\
&+ Lookahead (every 4 tok.) & 0.733 & 0.487   & 0.920 &  0.097 \\
&+ Lookahead (every 2 tok.) & 0.756 & 0.514  & 0.925 &  0.101 \\
&+ Lookahead (every tok.) & 0.767 & 0.524  & 0.926 & 0.102   \\
&+ CAD & 0.694 & 0.383  & 0.919 & 0.094  \\
&+ \ours & 0.752 & 0.504  & 0.920 & 0.096 \\
&+ \ours-d & 0.791 &0.523  & 0.921 & 0.104 \\
&+ \ours + CAD & 0.707 & 0.398  & 0.919 & 0.094 \\
&+ \ours-d + CAD & 0.735 &0.414  & 0.923 & 0.103\\
\cline{2-6}
&Nucleus & 0.545 & 0.364  & 0.902 & 0.082 \\
&+ CAD & 0.621 & 0.317  & 0.911 & 0.088 \\
&+ \ours & 0.730 & 0.489  & 0.917 & 0.093 \\
&+ \ours-d & 0.772 & 0.499  & 0.920 & 0.101 \\
&+ \ours + CAD & 0.695 & 0.373  & 0.918 & 0.093 \\
&+ \ours-d + CAD & 0.728 & 0.392  & 0.922 & 0.102 \\
\hline
\multirow{16}{*}{\rotatebox{90}{CNN/DM}} & Greedy & 0.840 & 0.506  & 0.922  & 0.146  \\
&+ Lookahead (every 8 tok.) & 0.843 & 0.511  & 0.923 &  0.147 \\
&+ Lookahead (every 4 tok.) & 0.848 & 0.514  & 0.925 &  0.149 \\
&+ Lookahead (every 2 tok.) & 0.866 & 0.546  & 0.930 & 0.157 \\
&+ Lookahead (every tok.) & 0.874 & 0.561  & 0.932 & 0.162  \\
&+ CAD & 0.828 & 0.301  & 0.917 & 0.173 \\
&+ \ours & 0.869 & 0.554  & 0.924 & 0.149 \\
&+ \ours-d & 0.884 & 0.570  & 0.925 & 0.151 \\ %
&+ \ours + CAD & 0.836 & 0.312  & 0.918 & 0.174 \\
&+ \ours-d + CAD & 0.849 & 0.330  & 0.919 & 0.178 \\
\cline{2-6}
&Nucleus & 0.706 & 0.310  & 0.907 & 0.122 \\
&+ CAD &0.777 & 0.232  & 0.911 & 0.157\\
&+ \ours & 0.857 & 0.521  & 0.922 & 0.142\\
&+ \ours-d & 0.872 & 0.533  & 0.923 & 0.145\\
&+ \ours + CAD & 0.828 & 0.291  & 0.916 & 0.169\\
&+ \ours-d + CAD & 0.841 & 0.313  & 0.918 & 0.174\\
\bottomrule
\end{tabular}%
}
\end{center}
\end{table*}

\begin{table*}[ht!]
\caption{\textbf{Faithfulness of the summaries generated with various decoding methods using \llama.} All the metrics are computed between the context document and the generated summary; higher is better.}
\label{tab:llama}
\begin{center}
\resizebox{0.6\linewidth}{!}{%
\begin{tabular}{c|lcccc}
\toprule
\textbf{} & \textbf{Method} &\textbf{AlignScore$\uparrow$} & \textbf{FactCC$\uparrow$}  & \textbf{BS-Fact$\uparrow$} & \textbf{Rouge-L$\uparrow$}   \\
\midrule
\multirow{4}{*}{\rotatebox{90}{\newsroom}} &Greedy & 0.701  & 0.321  & 0.897 & 0.161 \\
&+ CAD & 0.706 & 0.247 & 0.910 & 0.170 \\
&+ \ours & 0.715 & 0.328  & 0.906 & 0.162  \\
&+ \ours-d & 0.729 & 0.335  & 0.906 & 0.164 \\

\hline
\multirow{4}{*}{\rotatebox{90}{\xsum}} & Greedy & 0.798 & 0.406  & 0.931 & 0.221  \\
&+ CAD & 0.783 &0.335 & 0.931 & 0.237\\
&+ \ours & 0.800 & 0.410 & 0.932  & 0.221 \\
&+ \ours-d & 0.805 & 0.418   & 0.933 & 0.223\\
\hline
\multirow{4}{*}{\rotatebox{90}{CNN/DM }} & Greedy & 0.750 & 0.316  & 0.900 & 0.152  \\
&+ CAD &0.740 &0.251  & 0.919 & 0.176 \\
&+ \ours & 0.753 &  0.323  & 0.902 &  0.153 \\
&+ \ours-d &0.759 &0.327  & 0.902 &  0.154\\ 
\bottomrule
\end{tabular}%
}
\end{center}
\end{table*}

\subsection{Datasets and Models}
We consider two models: \flan XL~\citep{flant5} and \llama~\citep{llama}. We use the pretrained models without any further finetuning on individual datasets. 

We consider three datasets: \newsroom~\citep{newsroom}, \cnn~\citep{cnndm}, and \xsum~\citep{xsum}. We report numbers on the full test set of \cnn and \xsum, and randomly sample a subset of size 5000 from the \newsroom test set. The \xsum dataset uses the first sentence of the original article as the ground truth summary, and the rest of the article as the context document. Consequently, core information is sometimes missing from the context. To improve the completeness of the context and enable more meaningful comparison with the ground truth, we adopt a similar approach as \citet{wang2020asking} and prepend the ground truth summary back to the articles before performing summarization.

\subsection{Baselines and Implementation Details}
We examine four baseline decoding methods: greedy decoding, nucleus sampling, Lookahead~\citep{lookahead} (see \autoref{sec:related}), and CAD~\citep{shi2023trusting}. Note that Lookahead takes a long time to roll out future summaries and compute BS-Fact for each of the rollouts (for instance, generating 5000 Newsroom samples takes 108 hours). One natural way of increasing the speeding of this method is to perform ``lookahead" once every $l$ tokens instead of after every token. Thus, we consider four choices of $l$ for Lookahead: $l=1$ (the original version), $l=2$, $l=4$ and $l=8$. Additional implementation details can be found in \autoref{sec:more_impl} in the Appendix. 

We consider two versions of \ours: (1) \ours that only uses the conditional word probabilities for detection, which we refer as \ours in the tables; (2) \ours that uses both the conditional word probability and the token-context distance, which we refer as \ours-d. We use probability threshold $\delta=0.2$ and distance threshold $\varphi=0.5$ for \flan, and $\delta=0.3$ and $\varphi=0.9$ for \llama.

We evaluate \ours's performance with greedy decoding and nucleus sampling. Since \ours is complementary to most decoding methods, we can also use \ours in conjunction with some of the baselines. We report results of using \ours and CAD together. We do not evaluate using \ours with Lookahead due to the high computational cost. 

\subsection{Metrics}
To evaluate faithfulness, we compare the generated summaries with their \textit{source documents}. We use the following metrics: \textbf{AlignScore}~\citep{alignscore} and \textbf{FactCC}~\citep{factcc}, both of which employ learned models to score faithfulness; \textbf{BS-Fact}, which measures the BERTScore~\citep{bertscore} precision of a generated summary with respect to its context document; \textbf{ROUGE-L}~\citep{rouge}, which measures the longest common subsequence between the generation and reference. These metrics align relatively well with human judgement~\citep{frank} and have reasonable runtime.

We also report standard summarization metrics, including \textbf{ROUGE-L}, \textbf{BERTScore F1} and \textbf{Bleurt}~\citep{bleurt}, computed between the generated summaries and the datasets' \textit{ground truth summaries}. It should be noted that the models are used in a zero-shot manner. The quality of the generated summaries depends on the model's capabilities, and they may have different styles compared to the ground truth. Therefore, this comparison may not always yield informative results.

\subsection{Results}
We report the faithfulness performance of \flan on the different datasets in \autoref{tab:flan}, and the performance of \llama in \autoref{tab:llama}. Note that all metrics are computed between the source document and the generated summary. We report the metrics between the generated and ground truth summaries in \autoref{tab:flan_ref} and \autoref{tab:llama_ref} in the Appendix. For \flan, both Greedy with \ours and Lookahead at every token are competitive across datasets and metrics. Lookahead is slightly better according to BS-Fact and ROUGE-L, but is significantly slower as seen in \autoref{fig:time}. Greedy with \ours is comparable to Lookahead every 4 tokens and is still much faster. For \llama, \ours also attains performance gain. The improvement is smaller as \llama produces more faithful summaries than \flan. It is important to note that the absolute values of FactCC is smaller for \llama, because \llama produces much longer summaries than \flan, while FactCC has negative correlation with the summary length. We report the distribution of generated summary length in \autoref{fig:gen_len} in the Appendix, to show that the performance gain is not caused by producing shorter summaries. 

\begin{figure*}[ht!]
\vspace{-2ex}
    \centering
    \includegraphics[width=\linewidth]{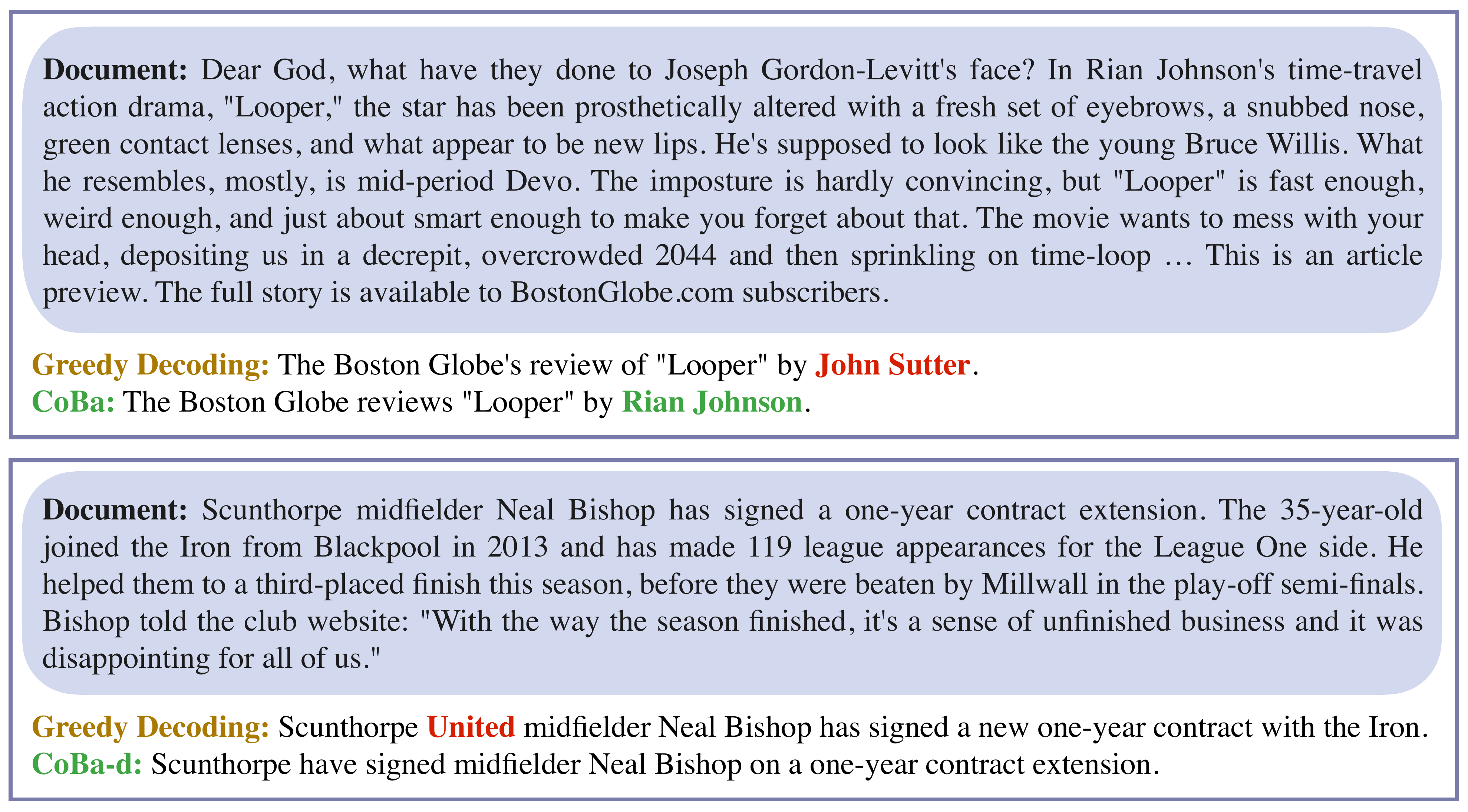}
    \caption{\textbf{Qualitative examples of greedy decoding vs. \ours and \ours-d.} The hallucinated content is marked in \textcolor{red}{red} and the corrected details are marked in \textcolor{teal}{green}. \ours and \ours-d correctly remove the hallucinated content by triggering backtracking at corresponding positions and generate summaries with more and faithful details.} 
    \label{fig:qualitative}
\end{figure*}

In \autoref{fig:qualitative}, we present two qualitative examples comparing greedy decoding vs. \ours and \ours-d. In the first example, the greedy decoding produces the summary "The Boston Globe's review of "Looper" by John Sutter." with a name that does not appear in the source document. Backtracking successfully replaces it with the correct name. 
In the second example, although the extended name of the soccer club can include "United" based on real world knowledge, the document itself only refers the soccer club as "Scunthorpe". \ours-d is able to detect this and remove "United". 

\subsection{Analysis}

\begin{figure}[ht!]
    \captionsetup{font=small}
    \centering
    \adjustbox{valign=t}{
    \begin{minipage}{\linewidth}
    \includegraphics[width=\linewidth]{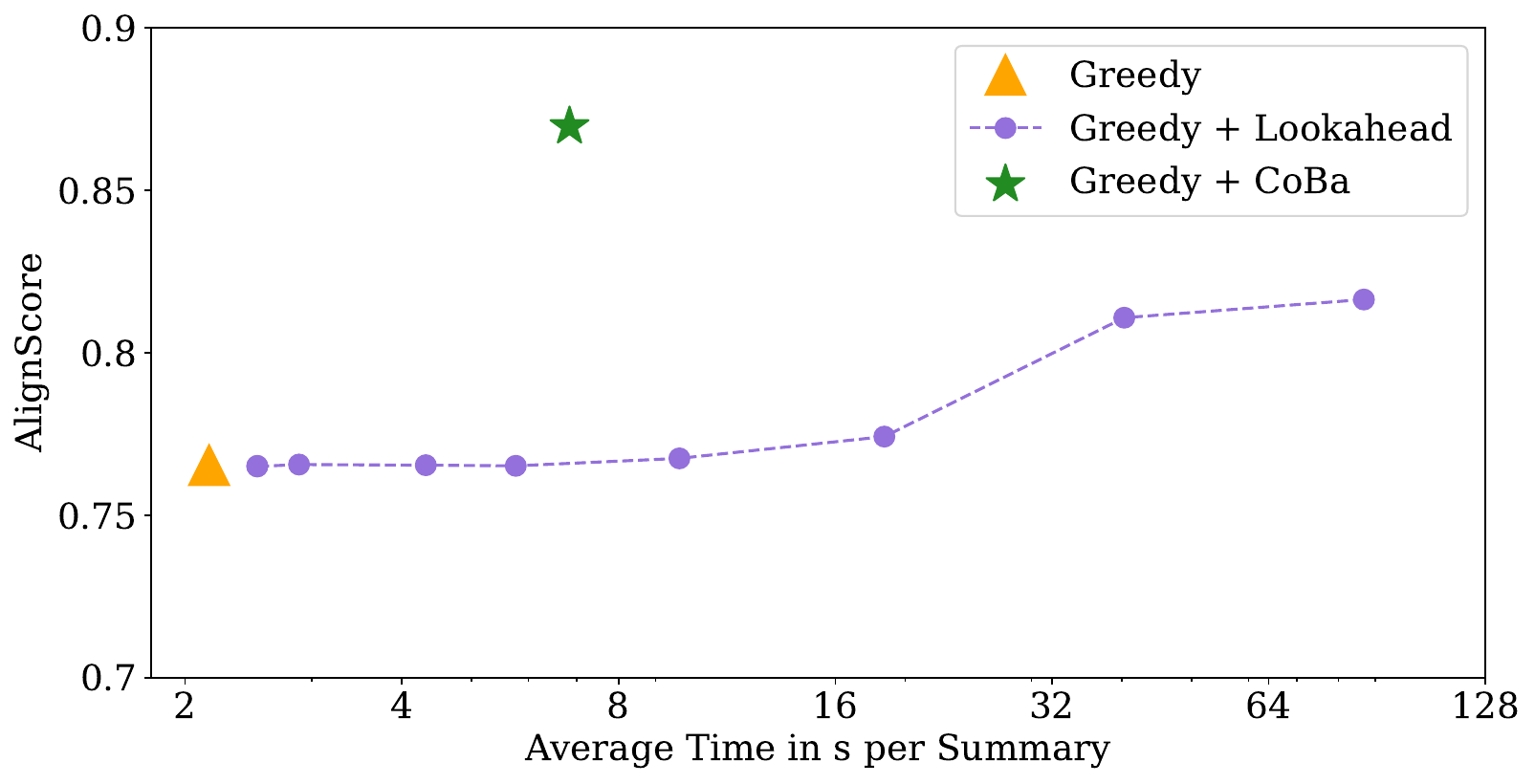}
  \caption{\textbf{AlignScore vs. Generation Time.} Note that the the x-axis is in log scale. The curve for Lookahead represents doing lookahead every $k$ tokens for $k$ from 200 to 1. \ours attains the highest AlignScore with more than 10x speedup.}
  \label{fig:time}
    \end{minipage}}
    \hfill
    \adjustbox{valign=t}{
    \begin{minipage}{\linewidth}
        \centering
\includegraphics[width=\linewidth]{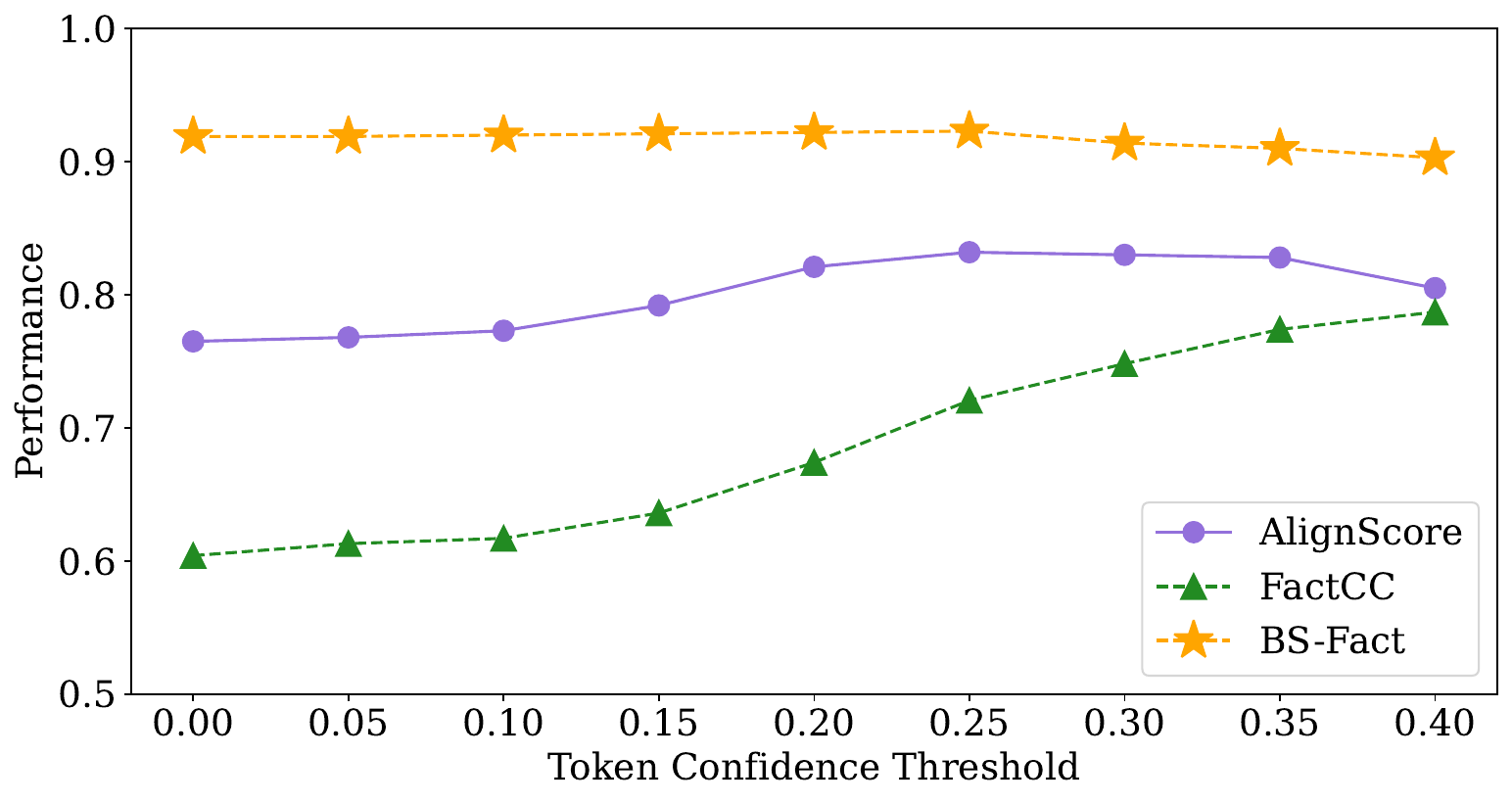}
\caption{\textbf{Ablation on the token confidence threshold for \ours.} High is better for all metrics. Most metrics saturate around threshold 0.2-0.25.}
\label{fig:confidence_abl}
    \end{minipage}}
\end{figure}

\textbf{Token Probability Threshold}. We examine the effects of using different values for the token confidence threshold, and present the results in \autoref{fig:confidence_abl}. We use the newsroom dataset and the \flan XL model. To better capture faithfulness, all the metrics are computed between source document and the generated summary. High value is better for all metrics. For AlignScore and BS-Fact, the improvement saturates at threshold 0.2-0.25, while FactCC continue to improve. 

\textbf{Embedding Distance Threshold}. We perform ablation studies on the choice of embedding distance threshold. Intuitively, the smaller the distance threshold is, the more similar the generated summaries are to their original documents. Results are presented in \autoref{tab:dist_abl}. "N/A" represents not applying the embedding distance threshold. We use token probability threshold 0.2, the \newsroom dataset, and the \flan XL model for the ablation experiments. Decreasing the threshold improves the performance, and the improvement saturates around threshold 0.5.

\begin{table}[ht!]
\caption{\textbf{Ablation on the threshold on token embedding distance.} We use token confidence threshold $\delta=0.2$ while varying the distance threshold $\varphi$ for all the experiments in this table. }
\label{tab:dist_abl}
\begin{center}
\resizebox{\columnwidth}{!}{%
\begin{tabular}{ccccc}
\toprule
\textbf{Dist. Thresh} & \textbf{AlignScore$\uparrow$} & \textbf{FactCC$\uparrow$}  & \textbf{BS-Fact$\uparrow$} & \textbf{Rouge-L$\uparrow$}  \\
\midrule
N/A & 0.821 & 0.674  & 0.923 & 0.138  \\
0.9 & 0.825 & 0.677  & 0.924 & 0.139 \\
0.7 & 0.859 & 0.699  & 0.925 & 0.143 \\
0.5 & 0.865 & 0.709  & 0.926 & 0.145 \\
0.3 & 0.867 & 0.718  & 0.920 & 0.146 \\
0.1 & 0.867 & 0.720  & 0.920 & 0.146 \\
\bottomrule
\end{tabular}%
}
\end{center}
\end{table}

\section{Limitations and Future Work}
In this study, we propose a method for reducing hallucinations in text summarization by backtracking. Our method consists of two steps: detection and backtracking. We employ two token level conditional probabilities and distance between generated tokens and context tokens to detect hallucinations. Both of these are effective ways of detecting hallucinated text, but there could be other complementary metrics that could improve detection. We defer the exploration of alternative metrics to future research endeavors.

While our primary focus in this paper is summarization models, our method can easily be extended to other applications where generating factual text is paramount. For instance, in question-answering systems which first retrieve relevant documents and then generate an answer, we can define the retrieved documents to be the context and employ \ours to produce factually correct answers.

\section{Conclusion}
Current decoding methods don't explicitly allow a model to re-generate some part of the generated  text when there is no highly probable completion to the partial text. Such a scenario would lead to hallucinations because the model is uncertain about how to complete the sentence and will sample a low probability word. 
We show that there is a relatively simple solution to mitigate hallucination, which we refer to as \ourslong{} (\ours). \ours{} is an inference-time method that requires no additional models, is computationally efficient, and can be directly applied to diverse summarization models without retraining. \ours detects hallucinations by using conditional probabilities of the generated tokens and measuring the distance between the generated text and the context. 
To correct the hallucinated text, it applies  
backtracking to before the hallucination and re-generates text to avoid ending up in positions with only low scoring token options.  
We empirically verify that \ours is able to identify and rectify hallucinated tokens during autoregressive decoding, and we show that \ours produces more factual summaries for various datasets. Our future work includes exploring other detection strategies as well as extending \ours to more diverse tasks. 
\section*{Acknowledgement}
This research is supported by a gift from the Simons Foundation, grants from the National Science
Foundation NSF (IIS-2107161, III1526012, IIS-1149882, OAC-2118310), Natural Sciences and Engineering Research Council of Canada (NSERC 567916)
and IIS-1724282), the Cornell Center for Materials
Research with funding from the NSF MRSEC program
(DMR1719875), LinkedIn and NewYork-Presbyterian Hospital.

\bibliography{reference}
\clearpage
\appendix

\onecolumn
\section{Appendix}
\label{sec:appendix}
\subsection{Additional Implementation Details}\label{sec:more_impl}
For the baselines, we use the Hugging Face implementation\footnote{\url{https://huggingface.co}} for greedy decoding and nucleus sampling, and the official code of lookahead\footnote{\url{https://github.com/amazon-science/faithful-summarization-generation}}. We use our own implementation for CAD as we did not find existing publicly available implementation. We use top-$p=0.9$ for nucleus sampling. Lookahead performs rollout for the top-$k$ tokens with the highest probabilities; we use $k=5$ following the original paper. CAD uses a scaling factor $\alpha$ when adjusting the conditional probabilities with the unconditional probabilities; we use $\alpha=0.5$ following the original paper. During generation, we set the minimum generation length to be 2 and maximum generation length to be 200 for all decoding methods.

\begin{table*}[bp!]
\caption{\textbf{Summarization metrics between the ground truth summaries from the dataset and the generated summaries using \flan.} Higher is better.}
\label{tab:flan_ref}
\begin{center}
\resizebox{0.7\linewidth}{!}{%
\begin{tabular}{c|lccc}
\toprule
\textbf{} & \textbf{Method} & \textbf{ROUGE-L$\uparrow$} &\textbf{BERTScore F1$\uparrow$} & \textbf{Bleurt$\uparrow$}  \\
\midrule
\multirow{16}{*}{\rotatebox{90}{\newsroom}} &Greedy & 0.312 & 0.890 & 0.441 \\
&+ Lookahead (every 8 tok.) & 0.313 & 0.890 & 0.441  \\
&+ Lookahead (every 4 tok.) & 0.314 & 0.891 & 0.442 \\
&+ Lookahead (every 2 tok.) & 0.323 & 0.892 & 0.451 \\
&+ Lookahead (every tok.) & 0.322 & 0.892 & 0.451  \\
&+ CAD & 0.281 & 0.883 & 0.412 \\
&+ \ours & 0.313 & 0.889 & 0.436  \\
&+ \ours-d & 0.306 & 0.885 & 0.428 \\
&+ \ours + CAD & 0.281 & 0.883 & 0.412 \\
&+ \ours-d + CAD & 0.267 & 0.878 & 0.399 \\
\cline{2-5}
&Nucleus & 0.267 & 0.883 & 0.406 \\
&+ CAD & 0.270 & 0.882 & 0.404  \\
&+ \ours & 0.306 & 0.888 & 0.432 \\
&+ \ours-d & 0.299 & 0.882 & 0.423 \\
&+ \ours + CAD & 0.282 & 0.883 & 0.412 \\
&+ \ours-d + CAD & 0.269 & 0.879 & 0.401  \\
\hline
\multirow{16}{*}{\rotatebox{90}{\xsum}} & Greedy & 0.422 & 0.920 & 0.540 \\
&+ Lookahead (every 8 tok.) & 0.426 & 0.921 & 0.543 \\
&+ Lookahead (every 4 tok.) & 0.431 & 0.922 & 0.546 \\
&+ Lookahead (every 2 tok.) & 0.467 & 0.927 & 0.569 \\
&+ Lookahead (every tok.) & 0.483 & 0.929 & 0.578 \\
&+ CAD & 0.399 & 0.916 & 0.522 \\
&+ \ours & 0.431 & 0.920 & 0.541 \\
&+ \ours-d & 0.462 & 0.919 & 0.548 \\
&+ \ours + CAD & 0.402 & 0.916 & 0.523 \\
&+ \ours-d + CAD & 0.440 & 0.919 & 0.535 \\
\cline{2-5}
&Nucleus & 0.297 & 0.902 & 0.459 \\
&+ CAD & 0.335 & 0.907 & 0.485\\
&+ \ours & 0.411 & 0.918 & 0.527\\
&+ \ours-d & 0.442 & 0.918 & 0.534\\
&+ \ours + CAD & 0.388 & 0.915 & 0.515\\
&+ \ours-d + CAD & 0.426 & 0.917 & 0.526\\
\hline
\multirow{16}{*}{\rotatebox{90}{CNN/DM}} & Greedy & 0.260 & 0.874 & 0.388\\
&+ Lookahead (every 8 tok.) & 0.261 & 0.875 & 0.389\\
&+ Lookahead (every 4 tok.) & 0.262 & 0.875 & 0.389\\
&+ Lookahead (every 2 tok.) & 0.265 & 0.876 & 0.395\\
&+ Lookahead (every tok.) & 0.265 & 0.876 & 0.396\\
&+ CAD & 0.248 & 0.871 & 0.392\\
&+ \ours & 0.256 & 0.873 & 0.382\\
&+ \ours-d & 0.256 & 0.872 & 0.383\\ 
&+ \ours + CAD & 0.248 & 0.871 & 0.391\\
&+ \ours-d + CAD & 0.246 & 0.870 & 0.389\\
\cline{2-5}
&Nucleus & 0.235 & 0.870 & 0.370\\
&+ CAD & 0.241 & 0.869 & 0.386\\
&+ \ours & 0.253 & 0.872 & 0.379\\
&+ \ours-d & 0.254 & 0.871 & 0.379\\
&+ \ours + CAD & 0.246 & 0.871 & 0.389\\
&+ \ours-d + CAD & 0.245 & 0.870 & 0.388\\
\bottomrule
\end{tabular}
}
\end{center}
\end{table*}
\begin{table*}[ht!]
\caption{\textbf{Summarization metrics between the ground truth summaries from the dataset and the generated summaries using \llama.} Higher is better.}
\label{tab:llama_ref}
\begin{center}
\resizebox{0.6\linewidth}{!}{%
\begin{tabular}{c|lccc}
\toprule
\textbf{} & \textbf{Method} & \textbf{ROUGE-L$\uparrow$} &\textbf{BERTScore F1$\uparrow$} & \textbf{Bleurt$\uparrow$}  \\
\midrule
\multirow{4}{*}{\rotatebox{90}{Newsroom}} &Greedy & 0.210 & 0.861 & 0.438 \\
&+ CAD & 0.207 & 0.872 & 0.429 \\
&+ \ours & 0.212 & 0.870 & 0.439 \\
&+ \ours-d & 0.214 & 0.868 & 0.436 \\
\hline
\multirow{4}{*}{\rotatebox{90}{\xsum}} & Greedy & 0.376 & 0.915 & 0.564 \\
&+ CAD & 0.362 & 0.908 & 0.533 \\
&+ \ours & 0.379 & 0.915 & 0.564 \\
&+ \ours-d & 0.383 & 0.915 & 0.564 \\
\hline
\multirow{4}{*}{\rotatebox{90}{CNN/DM }} & Greedy & 0.239 & 0.859 & 0.406 \\
&+ CAD & 0.236 & 0.872 & 0.401 \\
&+ \ours & 0.240 & 0.861 & 0.407 \\
&+ \ours-d & 0.240 & 0.860 & 0.405 \\ 
\bottomrule
\end{tabular}
}
\end{center}
\end{table*}
\subsection{Summarization Metrics between Ground Truth and Generated Summaries}
We report the summarization metrics between the ground truth summaries and the generated summaries from \flan in \autoref{tab:flan_ref} and \llama in \autoref{tab:llama_ref}. All decoding methods with both models demonstrate reasonable performance. It is important to note that the ground truth summaries for each dataset are collected by distinct criteria: CNN/Dailymail~\citep{cnndm} uses the human-written story highlights in bullet points, XSUM takes the first sentence of a document~\citep{xsum, wang2020asking}, and Newsroom uses the HTML metadata~\citep{newsroom}. As the models are not further finetuned on the individual datasets, their summaries often exhibit different styles from the ground truth summaries. Consequently, the summarization metrics only provide limited insights into the quality of the generated summaries.

\subsection{Generated Summary Lengths}
\begin{figure*}[h]
    \captionsetup{font=small}
    \adjustbox{valign=t}{
    \begin{minipage}{0.49\linewidth}    \includegraphics[width=\linewidth]{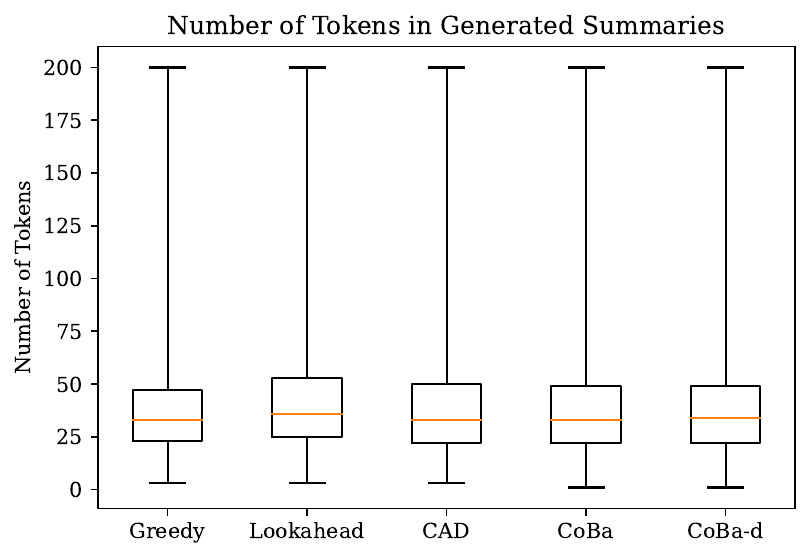}
    \end{minipage}}
    \adjustbox{valign=t}{
    \begin{minipage}{0.49\linewidth}
        \centering
\includegraphics[width=\linewidth]{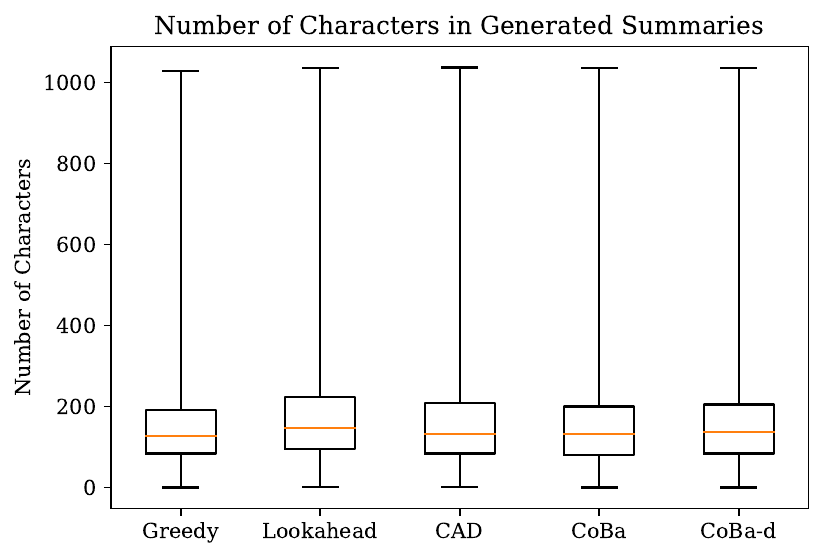}
    \end{minipage}}
    \caption{\textbf{Number of tokens (left) and number of characters (right) in the generated summaries from \flan on the Newsroom dataset.} The lengths have similar distributions across generation methods.}\label{fig:gen_len}
\end{figure*}

\autoref{fig:gen_len} shows the lengths of generated summaries from \flan on the Newsroom dataset. In general, the length distribution is similar across different decoding methods.

\end{document}